\documentclass[conference]{IEEEtran}
\IEEEoverridecommandlockouts
\usepackage{cite}
\usepackage{amsmath,amssymb,amsfonts}
\usepackage{algorithmic}
\usepackage{graphicx}
\usepackage{textcomp}
\usepackage{enumitem}
\usepackage{xcolor}
\usepackage[ruled,vlined]{algorithm2e}
\usepackage{subfig}
\usepackage{multirow}
\usepackage{url}
\usepackage{bm}
\usepackage{hyperref}
\usepackage{multirow}
\usepackage{booktabs}

\def\BibTeX{{\rm B\kern-.05em{\sc i\kern-.025em b}\kern-.08em
    T\kern-.1667em\lower.7ex\hbox{E}\kern-.125emX}}
\begin{document}

\title{Harnessing Hypergraphs in Geometric Deep Learning for 3D RNA Inverse Folding }

\author{
\IEEEauthorblockN{Guang Yang\textsuperscript{1}, Lei Fan\textsuperscript{2}\IEEEauthorrefmark{2}}
\IEEEauthorblockA{\textsuperscript{1}\textit{Faculty of Computing}, \textit{Harbin Institute of Technology}, Harbin, China}
\IEEEauthorblockA{\textsuperscript{2}\textit{Computer Science and Engineering, UNSW}, Sydney, Australia}
\IEEEauthorblockA{\IEEEauthorrefmark{2}Corresponding author, Email: lei.fan1@unsw.edu.au}
}

\def\etal{\emph{et al. }}
\def\ie{\emph{i.e., }}
\def\eg{\emph{e.g., }}

\maketitle
\begin{abstract}
The RNA inverse folding problem, a key challenge in RNA design, involves identifying nucleotide sequences that can fold into desired secondary structures, which are critical for ensuring molecular stability and function. The inherent complexity of this task stems from the intricate relationship between sequence and structure, making it particularly challenging. 

In this paper, we propose a framework, named HyperRNA, a generative model with an encoder-decoder architecture that leverages hypergraphs to design RNA sequences. Specifically, our HyperRNA model consists of three main components: preprocessing, encoding and decoding. 
In the preprocessing stage, graph structures are constructed by extracting the atom coordinates of RNA backbone based on 3-bead coarse-grained representation. The encoding stage processes these graphs, capturing higher order dependencies and complex biomolecular interactions using an attention embedding module and a hypergraph-based encoder. Finally, the decoding stage generates the RNA sequence in an autoregressive manner.
We conducted quantitative and qualitative experiments on the PDBBind and RNAsolo datasets to evaluate the inverse folding task for RNA sequence generation and RNA-protein complex sequence generation. The experimental results demonstrate that HyperRNA not only outperforms existing RNA design methods but also highlights the potential of leveraging hypergraphs in RNA engineering.

\end{abstract}


\section{Introduction}
RNA design is a specialized field within bioinformatics and synthetic biology focused on creating RNA sequences capable of folding into specific spatial structures or performing particular functions. 
RNA molecules are involved in various cellular processes, such as serving as messengers for genetic information (mRNA), forming the core components of ribosomes (rRNA) and fulfilling regulatory and catalytic roles (\eg tRNA, miRNA, ribozymes). 
Designing RNA sequences holds significant potential for applications in medicine biotechnology and long-term research~\cite{b1,b2,b3}.

%waiting
RNA design typically involves the primary, secondary and tertiary structures, encompassing genetic information, structural formations and overall functionality.
The RNA inverse folding problem refers to the challenge of identifying an RNA sequence that can fold into a desired secondary structure, which is essential for the molecular stability, function and interactions.
This process requires determining a nucleotide sequence capable of folding into a specified RNA secondary structure under physiological conditions, such as temperature.
By solving the inverse folding problem, researchers can create RNA sequences with desired properties and functions, enabling advancements in areas such as gene regulation, drug discovery and disease treatment~\cite{b4,b5}.

\begin{figure}[!t]
    \centering
    \includegraphics[width=1\linewidth]{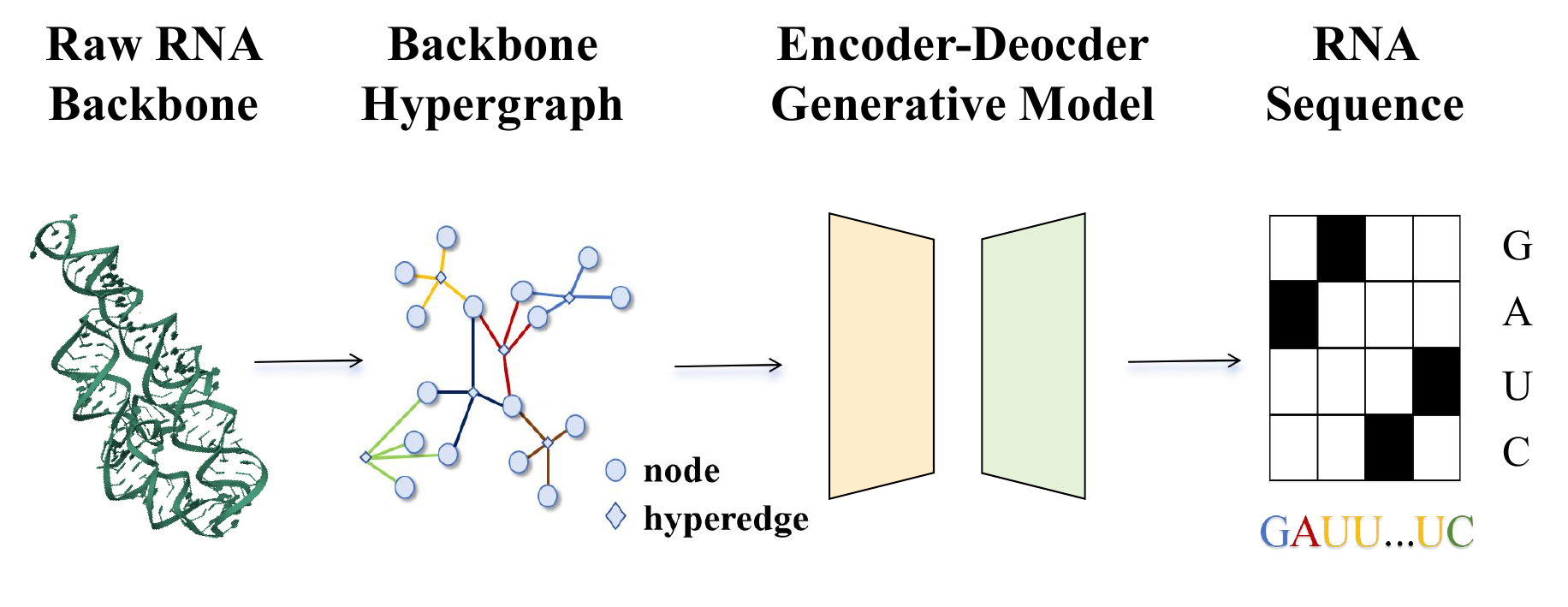}
    \caption{Our framework is divided into three steps: preprocessing the RNA backbone, encoding graph features and generating RNA sequences. }
    \label{fig:summarize}
\end{figure}

The primary challenges in RNA design arise from the limited availability of raw RNA data and the intricate structure of RNA.
Compared to protein designs that rely on stable tertiary structures and predictable folding patterns~\cite{Kuhlman2019,Huang2016}, RNA exhibits greater conformational flexibility, influenced by base pairing interactions between chains and base stacking between adjacent nucleotide rings~\cite{challenge}. On the other hand, the nucleotides in the RNA backbone contain a wider variety of atoms, requiring the parameterization of most positions by torsion angles. This complexity further complicates the modeling of the RNA backbone.

Recently, with the deep learning techniques~\cite{fan2024patch,tang2025prototype,fan2025grainbrain}, several studies based on graph learning have been developed for RNA design. For example, Joshi~\etal\cite{RD8} first attempted to solve the 3D RNA inverse design problem by introducing a geometric learning pipeline.
Similarly, RNAFlow~\cite{rnaflow} and RNA-FRAMEFLOW~\cite{rna-frame-flow} are proposed based on flow matching models.
These methods convert and represent the RNA structure by constructing graphs with Graph Neural Networks (GNNs).
The coordinates and atoms are treated as nodes, capturing and aggregating their relationships to predict potential sequences. 
These approaches have demonstrated the potential of using graph structures combined with deep learning techniques, particularly generative models. 
However, their performance is limited by the use of vanilla graph feature extraction with basic GNNs, which struggle to represent inherent multi-atomic relationships effectively.

Hypergraph methods have emerged as powerful tools, gaining traction in various domains including social network analysis~\cite{hgusing1}, computer vision~\cite{HG2,HG3,M3Surv} and autonomous driving~\cite{hgusing2}.
Hypergraphs offer unique advantages in representing complex data structures and interactions. 
For example, Zhu~\etal\cite{hgusing1} utilized hypergraphs to model user-item-user relationships, with hyperedges representing triplets, thereby improving social recommendation accuracy via hypergraph convolutional networks.
Similarly, Xu~\etal\cite{hgusing2} employed a multiscale hypergraph neural network to capture both pair-wise and group-wise interactions in trajectory prediction, achieving superior performance in relational reasoning and prediction accuracy.

Compared to traditional graphs, hypergraphs can capture higher-order relationships by connecting multiple nodes within hyperedges, allowing for a more accurate and comprehensive representation of intricate dependencies. 
This is particularly beneficial for RNA inverse folding problems, where the interactions between various nucleotides and structural elements are not simply pairwise but often involve multi-node dependencies. 
Hypergraphs are better suited to model the nonlinear, multi-way interactions between RNA bases, enabling the capture of subtle structural features and conformations that traditional graphs might miss. 

In this paper, we explore the use of hypergraphs for the 3D RNA inverse folding problem. 
We leverage hypergraphs to effectively handle the complex structure of RNA and further capture interactions between nucleotides, as shown in Fig.~\ref{fig:summarize}. 
We propose a framework, named HyperRNA, which consists of three steps: preprocessing the RNA backbone, encoding graph features and generating RNA sequences.
Specifically, we first extract the RNA backbone using a 3-bead coarse-grained representation~\cite{3bg}, where atoms are treated as nodes and their edges are established by leveraging the $k$-nearest neighbors (kNN) method. Graph node features are fed into a self-attention embedding module to capture intrinsic dependencies.  Then, we employ a hypergraph-based generative model with an encoder-decoder architecture to learn and sequentially generate nucleotides for RNA sequences. 

The contributions of our work can be summarized as below:
\begin{itemize}
    \item We propose a framework, named HyperRNA, leveraging a hypergraph-based generative model designed for the RNA inverse folding problem. 
    \item Our approach incorporates a self-attention embedding module to extract graph features and capture intrinsic dependencies.
    \item We conducted experiments on the PDBBind~\cite{pdbbind} and RNAsolo~\cite{Adamczyk2022} datasets, demonstrating consistent improvements across several metrics with our model. Additionally, we verified the effectiveness of incorporating hypergraphs into existing models~\cite{RD8}.
\end{itemize}

\section{Related Work}
\subsection{RNA Design}
\textbf{RNA structure design} can be categorized into protein-conditional and unconditional approaches.
Early protein-binding RNA design strategies relied on physical structure and molecular dynamics, such as selecting appropriate RNA from a large pool~\cite{RSD1} or using molecular dynamics simulations~\cite{RSD2,RSD3}. 
Recent advancements include methods based on Long Short-Term Memory (LSTM) networks~\cite{RSD4} and Variational Autoencoders (VAEs)~\cite{RSD5}. Nori~\etal\cite{rnaflow} proposed RNAFlow using a flow matching model for RNA sequence-structure design.
Unconditional RNA design using computational methods includes non-aligned algorithm-based structure generation~\cite{RD4} and generative models~\cite{rna-frame-flow}. The \textit{SE(3)}-discrete diffusion model (MMDiff) developed by Hoetzel~\etal\cite{RSD6} jointly generates nucleotide sequences and RNA structures.

\textbf{RNA inverse folding}  has lagged behind protein inverse folding studies \cite{De-novo}, primarily due to the limited amount of known RNA 3D structures. 
The application of deep learning to computational RNA design was initially explored by Schneider~\etal\cite{RSD1}. Most RNA design tools focus primarily on secondary structure without considering 3D geometry~\cite{RD2} and use non-learning algorithms to align 3D RNA fragments~\cite{RD3,RD4}. However, due to the flexible nature of RNA strands, there are multiple conformations, complicating the design process~\cite{RD5,RD6,RD7}.
Recently, gRNAde \cite{RD8}, a GNN-based approach for 3D RNA inverse folding, introduced a geometric learning pipeline utilizing an encoder-decoder architecture. 

In this work, we aim to explore the use of hypergraphs in an encoder-decoder generative model to extract and capture the intricate associations within 3D RNA structures.

\begin{figure*}[t]
    \centering
    \includegraphics[width=1\linewidth]{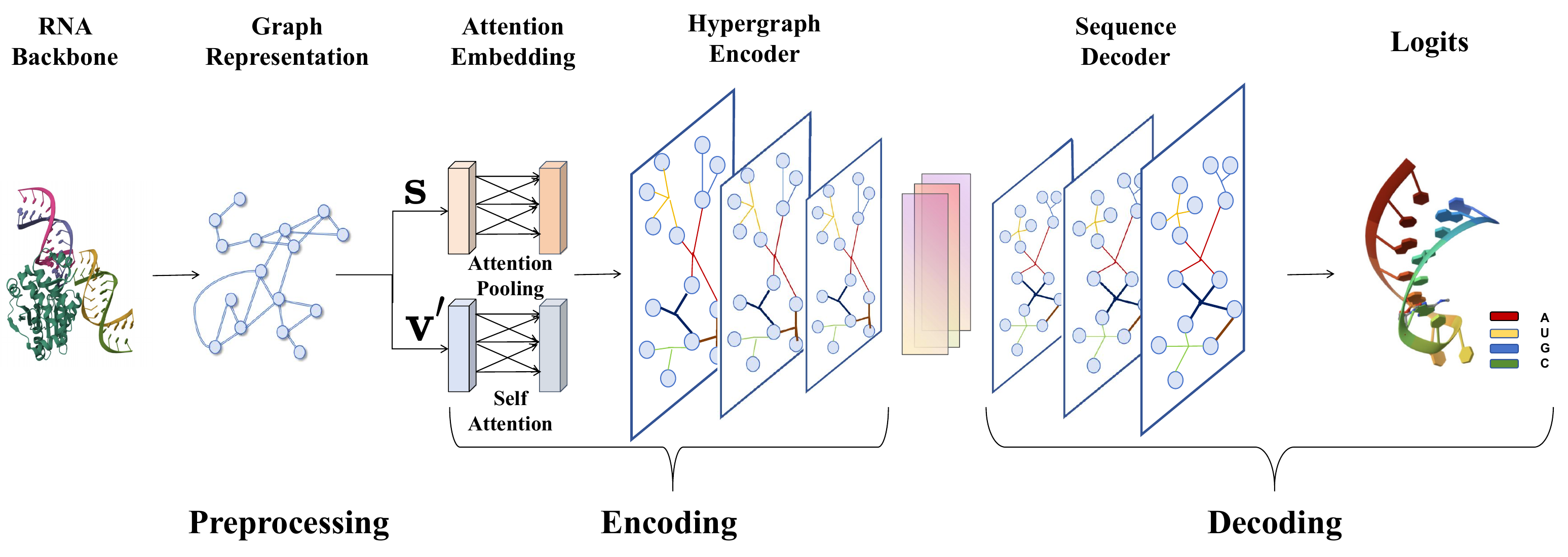}
    \caption{\textbf{HyperRNA model architecture.} It consists of three steps: preprocessing, encoding and decoding. First, atom coordinates of the RNA backbone are extracted using 3-bead coarse-grained representation to construct graph structures. These graphs are then embedded using the attention embedding module and processed by a hypergraph-based encoder. Finally, the hypergraph-based decoder utilizing the autoregressive model generates RNA sequences.}
    \label{fig:structure}
\end{figure*}

\subsection{Hypergraph Learning}
Hypergraph learning began as a propagation process on hypergraph structures, focusing on minimizing label differences between strongly connected vertices~\cite{qu2024boundary,qu2025multimodal,qu2025spatially}. For example, Huang~\etal\cite{HG2} used a hypergraph to represent the spatio-temporal relationships among patches, framing the extraction of prominent objects from a scene as a hypergraph cut problem.
Yu~\etal\cite{Yu2012} proposed an adaptive hypergraph learning method for transductive image classification, addressing the challenges of using hypergraphs for transductive image classification.
Due to their strong representational power, hypergraphs were used in multimodal fields, with multiple hyperedges representing different data types \cite{HG4,HG5,HG6}. Feng \etal\cite{HG7} combined hypergraphs with graph neural networks (GNNs), and introduced a message passing model for hypergraphs in data representation learning. 

Considering hypergraphs that capture high-order interactions for better modeling intricate relationships~\cite{jing2025multi,wang2025hypergraph}, we use them to represent RNA molecules and capture their biological characteristics, such as protein binding, RNA interactions and intermolecular interactions between RNA nucleotides.

\section{Methods}

\subsection{Preliminaries}
\textbf{RNA inverse folding definition.} Given a protein-binding RNA structure comprising both a protein and RNA, we utilize a 3-bead coarse-grained representation~\cite{3bg} to capture the essential atomic coordinates of backbone structures. 
This representation focuses on key atomic structures that define the core architecture, enabling an effective description of the spatial configuration. 

Specifically, a protein structure is denoted as $\mathcal{P} \in \mathbb{R}^{L_{P} \times 3 \times 3}$, where $L_{P}$ is the number of residues in the protein, and each protein residue is represented by the coordinates of three principal atoms: $N$ (nitrogen atom), $C_\alpha$ (alpha carbon atom) and $C$ (carbonyl carbon atom).
An RNA structure is denoted as $\mathcal{R} \in \mathbb{R}^{L_{R} \times 3 \times 3}$, where $L_{R}$ is the number of nucleotides in the RNA sequence, and each nucleotide is characterized by the coordinates of three principal atoms: $P$ (phosphorus atom), $C'_4$ ($4'$ carbon atom on the ribose ring) and $N_1/N_9$ (base nitrogen atom). 
The goal is to design a model that predicts the RNA sequence $\hat{R_s} \in \mathbb{R}^{L_R}$ (a string of $L_R$ nucleotides) based on the structural information of the RNA backbone, including atom coordinates $s$ and the true RNA sequence $R_s$.

\textbf{Hypergraph construction.}
A hypergraph is formally defined as $\mathcal{G} = \{\mathcal{V}, \mathcal{E}, \mathbf{W}\}$. Specifically, a set of vertices is defined as $\mathcal{V}=\{v_1,\cdots,v_{N_v}\}$, a set of hyperedges is defined as $\mathcal{E}=\{e_1, \cdots, e_{N_e}\}$ and a diagonal matrix of the hyperedge weights is denoted as $\mathbf{W} = [w(e_1),\cdots,w(e_{N_e})]$, where $N_v$ (\ie $|\mathcal{V}|$) and $N_e$ (\ie $|\mathcal{E}|$) indicate the number of vertices and hyperedges. A hypergraph $\mathcal{G}$ can be further described by an incidence matrix $\mathbf{H} \in \mathbb{R}^{|\mathcal{V}| \times |\mathcal{E}|}$ as follows:
\begin{equation}
\mathbf{H}(v,e)=\begin{cases}
    1 &\text{if }v\in e
    \\0 & \text{if }v\notin e
\end{cases},
\end{equation}
where $\mathbf{H}(v,e)$ denotes an entry of $\mathbf{H}$, indicating whether vertex $v$ is part of hyperedge $e$. 
The degree of a vertex $v \in \mathcal{V}$ is defined as:
\begin{equation}
d(v) = \sum_{e \in \mathcal{E}} w(e) * \mathbf{H}(v, e) ,   
\end{equation}
where $w(e)$ is the weight of the corresponding hyperedge $e$. The degree of a hyperedge $e \in \mathcal{E}$ is defined as $\delta(e) = \sum_{v \in \mathcal{V}} \mathbf{H}(v, e)$, representing the number of vertices incident on $e$. Then, the diagonal matrices $\mathbf{D_v} \in \mathbb{R}^{|\mathcal{V}| \times |\mathcal{V}|}$ and $\mathbf{D_e} \in \mathbb{R}^{|\mathcal{E}| \times |\mathcal{E}|}$ represent the vertex and hyperedge degrees, respectively.

\textbf{Hypergraph Neural Network (HGNN).} It processes hypergraphs by leveraging hyperedges that connect multiple nodes. It typically consists of an input layer for node features, hypergraph convolution layers for message passing, and an output layer for updated node representations. In an HGNN, messages are aggregated from nodes within the same hyperedge, and node features are updated based on these aggregated messages. Given a hypergraph signal $\mathbf{X} \in \mathbb{R}^{n\times d}$, where $n$ is the number of nodes and $d$ is the dimensionality of the node features, the hyperedge convolution operation can be formulated as follows:
\begin{align}
    \mathbf{Z}=\sigma(\mathbf{D_v}^{-\frac{1}{2}}\mathbf{H}\mathbf{W}\mathbf{D_e}^{-1}\mathbf{H}^{T}\mathbf{D_v}^{-\frac{1}{2}}\mathbf{X}\Theta),
\end{align}
where $\mathbf{Z} \in \mathbb{R}^{n\times d'}$ is the updated node feature matrix, $\Theta\in\mathbb{R}^{d\times d'}$ is a learnable weight matrix, and $\sigma$ is a nonlinear activation function. This process is iteratively refined through multiple layers, allowing the HGNN to capture complex, higher-order interactions and to improve its representation of intricate data structures.

\subsection{Overview}
Our proposed HyperRNA model mainly consists of three steps: preprocessing, encoding and decoding, as shown in Fig.~\ref{fig:structure}. Specifically, in the preprocessing stage, the RNA and protein backbone data are transformed into atom coordinates $\mathcal{R}$ and $\mathcal{P}$ using a 3-bead coarse-grained representation, preparing them for subsequent stages. $n$ atoms are treated as nodes to construct an adjacency matrix $\mathcal{A} \in \mathbb{R}^{N \times N}$ using the $k$-Nearest Neighbors (kNN) method as follows:
\begin{equation}
N_k(v_i) = \{v_j \mid \text{rank}(d(v_i, v_j)) \leq k, i\ne j \},    
\end{equation}
where $ d(v_i, v_j) $ represents the $L_2$ distance between nodes $v_i$ and $v_j$. The set $N_k(v_i)$ includes the ($k$) nearest neighbors of node $v_i$ based on the distance ranking. For each vertex $v_i$, if $v_j \in N_k(v_i)$, then $\mathcal{A}_{ij} = 1$; otherwise, $\mathcal{A}_{ij} = 0$.

These atom coordinates, $\mathcal{R}$ and $\mathcal{P}$, along with the adjacency matrix $\mathcal{A}$, are then featurized to generate a graph structure $\mathcal{G}=\{\mathcal{A}, \mathbf{s}, \mathbf{v}'\}$ using a geometric featurization method~\cite{gR31,gR32}. Here, $\mathbf{s} \in \mathbb{R}^{n \times d_e}$ represents scalar features, and $\mathbf{v}' \in \mathbb{R}^{n \times d_v \times 3}$ represents vector features, with $d_e$ and $d_v$ denoting the number of scalar and vector feature channels, respectively. 
These features include forward and reverse unit vectors along the backbone from the $5'$ end to the $3'$ end, as well as unit vectors, distances, angles and torsions from each \textit{$C4'$} atom to the $P$ and $N_1/N_9$ atoms.

We employ a hypergraph-based encoder-decoder architecture to process the featurized graph information and synthesize RNA sequences. The encoder maps the graph features into a latent representation, allowing the model to capture complex higher-order interactions inherent in the RNA structure. The decoder, using an autoregressive model, reconstructs the RNA sequence from the latent representation, ensuring alignment with the underlying structural and interaction patterns.

\subsection{Encoding and Decoding}

\textbf{Feature encoding.} This stage takes the RNA-protein complex backbone graph $\mathcal{G}$ as input, capturing high-order dependencies and complex biomolecular interactions to generate latent representations. The encoding stage comprises an attention embedding module and a hypergraph-based encoder. The attention embedding module is applied to the scalar features $\mathbf{s}$ and vector features $\mathbf{v}'$ in the graph $\mathcal{G}$, optimizing these features for hypergraph propagation. The hypergraph-based encoder then leverages the HGNN's message-passing mechanism to update node features, resulting in pooled latent representation $\mathbf{s}_p\in\mathbb{R}^{n\times d_e}$ and $\mathbf{v}'_p\in\mathbb{R}^{n\times d_v\times 3}$.

\begin{figure}[!t]
    \centering
    \includegraphics[width=1\linewidth]{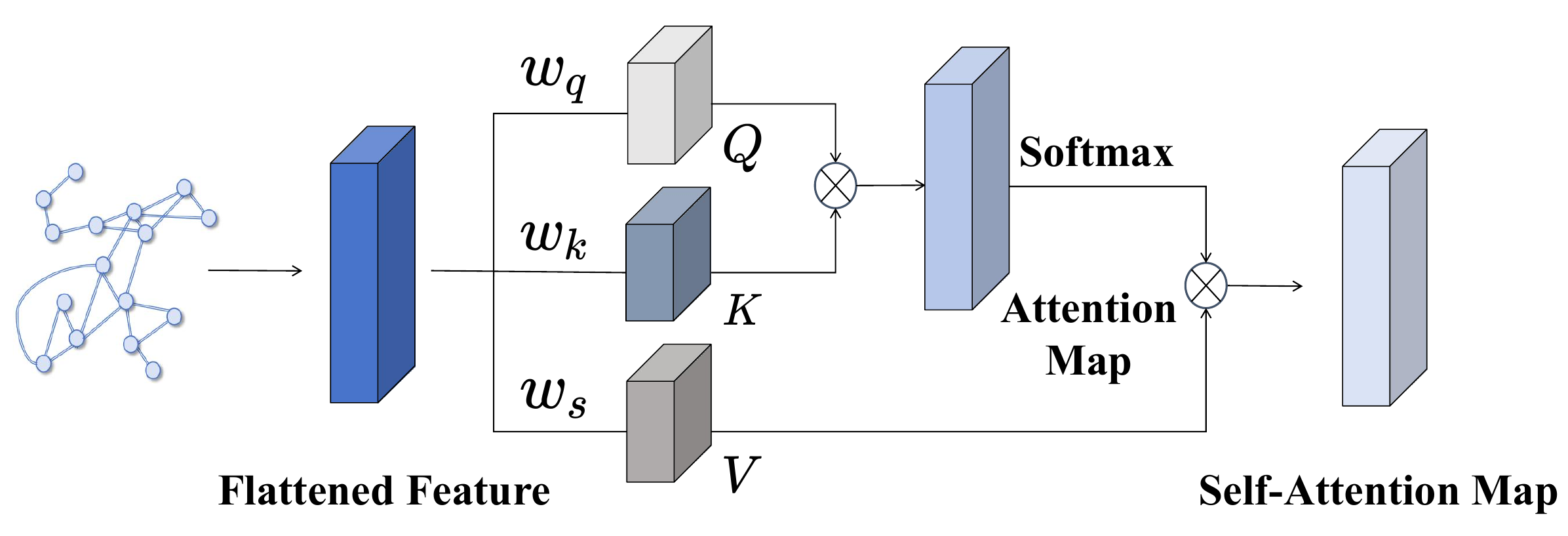}
    \caption{\textbf{The architecture of Self-Attention mechanism}. $Q,K$ and $V$ are calculated by using the graph vector feature with their corresponding learnable weight metrics $\mathbf{w}_q$, $\mathbf{w}_k$ and $\mathbf{w}_v$.}
    \label{fig:SA}
\end{figure}

Specifically, for the attention embedding module, a multi-head attention mechanism is applied to the flattened vector features $\mathbf{v}'$, as illustrated in Fig.~\ref{fig:SA}. It includes $h$ heads, and each head independently computes the scaled dot-product attention $\mathbf{v}_h^i$ ($i=\{1,\cdots, h\}$) as follows:
\begin{equation}
    \mathbf{v}_h^i = \text{softmax}\left(\frac{\mathbf{v}'\mathbf{w}^{i}_q{(\mathbf{v}'\mathbf{w}^{i}_k)}^T}{\sqrt{3\cdot d_v / h}}\right)\mathbf{v}'\mathbf{w}^{i}_v,
\end{equation}
where $\mathbf{w}^i_q$, $\mathbf{w}^i_k$ and $\mathbf{w}^i_v$ are learnable weight matrices for computing the query $Q$, key $K$ and value $V$ matrices respectively. The resulting output from the multi-head attention mechanism is denoted as $\mathbf{v}'_a\in\mathbb{R}^{n\times d_v\times 3}$.

The scalar features $\mathbf{s}$ are divided into five distinct types: forward radial basis values, backward radial basis values, carbon-nitrogen radial basis values, carbon-phosphorus radial basis values and token representations. For each type $\mathbf{s}_i$ ($i\in\{1,\cdots,5\}$), an attention pooling operation is employed to obtain the attention scores $\gamma_i$ as follows:
\begin{equation}
     \gamma_i=\frac{\exp({\mathbf{w}_p^i} \mathbf{s}_i)}{\sum_{j} \exp(\mathbf{w}_p^j \mathbf{s}_j)} ,
\end{equation}
where ${\mathbf{w}_p^i}$ is a learnable weight matrix. These scores are used to weight the scalar features, aggregating them into a pooled representation $\mathbf{s}_a=\sum_{i=1}^{5}{\gamma_i \mathbf{s}_i}$. The attention embedding module results in the embedded graph $\mathcal{G}_a = \{\mathcal{A}, \mathbf{s}_a, \mathbf{v}'_a\}$.

%  \in\mathbb{R}^{n\times d_e \times 3}$
%The resulting hypergraph adjacency matrix $\mathbf{A'} \in \mathbb{R}^{n \times n}$ represents the hyperedges, where $\mathbf{A'}_{i,j}$ is 1 if existing a hyperedge \(e \in \mathcal{E}\) such that both \(i \in e\) and \(j \in e\); otherwise, $\mathbf{A'}_{i,j}$ is 0.

For the hypergraph-based encoder, the process includes constructing the hypergraph and updating node features with $L$ HGNN layers. First, a hypergraph $\mathcal{G}'=\{\mathcal{V},\mathcal{E},\mathbf{W}\}$ is constructed from the embedded graph $\mathcal{G}_a$ using the graph adjacent matrix $\mathcal{A}$, where hyperedges $\mathcal{E}$ are formed based on node neighbor relationships scalar features $\mathbf{s}_{a}$ and vector features $\mathbf{v}'_a$. For each HGNN layer, information is gathered from the vertices to the hyperedges, and node features are updated based on the information propagated during the $i$-th ($i\in \{1,\cdots,L\}$) feature update process:
\begin{equation}
    \mathbf{s}^{(i+1)}_a,\mathbf{v}'^{(i+1)}_a= \sigma\left(\mathbf{{D_v}^{-1}\mathcal{A}D^{-1}_e{\mathcal{A}}^T}(\mathbf{s}^{(i)}_a,\mathbf{v}'^{(i)}_a) \Theta \right),
\end{equation} 
where $\mathbf{D_v}$ and $\mathbf{D_e}$ are the vertex degree matrix and hypergraph degree matrix respectively, $\Theta$ and $\sigma$ are the learnable weight matrix and a nonlinear activation function respectively.

The final updated node features $\mathbf{s}^{(L)}_a$ and $\mathbf{v}'^{(L)}_a$ are normalized into encoded features $\mathbf{s}_e$ and $\mathbf{v}'_e$. These encoded features are then pooled to obtain the latent representations $\mathbf{s}_p$ and $\mathbf{v}'_p$, which are used in the decoding stage.

\textbf{Sequence decoding.}
The decoder receives the pooled scalar features $\mathbf{s}_p$ and vector features $\mathbf{v}'_p$ to predict the nucleotide sequence using $L$ Geometric Vector Perceptron (GVP) layers~\cite{gR32}. The prediction is performed through an autoregressive process~\cite{teacher-forcing}, which sequentially generates the RNA sequence from the $5'$ to $3'$ end. 
At each step $t$, the decoder leverages GVPs to process the node features $\mathbf{s}_p$ and $\mathbf{v}'_p$, along with the previously generated nucleotides $\{n_{t-1}, n_{t-2}, \cdots, n_0\}$ to predict the next nucleotide $n_t$, as expressed:
\begin{equation}
    n_t = \text{softmax}\left\{\text{GVP}(\mathbf{s}_p, \mathbf{v}'_p, \{n_{t-1},\cdots, n_0\})/\tau\right\},
\end{equation}
where $\tau$ denotes the temperature that controls the entropy of the probability distribution.

For the obtained $n_t$, the decoder determines its nucleotide identity and classifies it into one of the four nucleotides: {A, G, C, U}.
The decoder iteratively applies this process across all timesteps $T$ to generate the predicted RNA sequence $\hat{R}_s$.

\subsection{Training and Inference}

\textbf{Training.} Our proposed HyperRNA model is trained using a self-supervised approach by minimizing the total loss $ \mathcal{L}_{total}$, which is defined as:
\begin{equation}
\begin{aligned}
\mathcal{L}_{total} &= \mathcal{L}_{seq} + \mathcal{L}_{str} \\
    &= -\sum_{i} R_{s_i} \log(\hat{R}_{s_i}) + \frac{1}{n} \sum_{j} (\hat{s}_j - s_j)^2.
\end{aligned}
\end{equation}
where $\mathcal{L}_{seq}$ and $\mathcal{L}_{str}$ represent the sequence loss and structure loss respectively. Specifically, $\mathcal{L}_{seq}$ leverages the cross-entropy (CE) objective to optimize the predicted RNA sequence by minimizing the difference between the true sequence $R_s$ and the predicted sequence $\hat{R_s}$. $\mathcal{L}_{str}$ uses the mean squared error (MSE) to ensure that the predicted RNA sequence accurately reflects the true 3D conformation by minimizing the discrepancy between the true backbone coordinates $s$ and the predicted coordinates $\hat{s}$, where $\hat{s}$ is derived from folding the predicted sequence $\hat{R}_s$ using the RF2NA method~\cite{rf2na}.
During training, we apply autoregressive teacher forcing~\cite{teacher-forcing} to ensure alignment with the ground-truth sequence by feeding the true base identity into the decoder at each step.

\textbf{Inference.} Since our model is designed for generating RNA sequences conditioned on the protein-binding RNA backbone, we integrate the Trajectory-to-Seq method \cite{rnaflow} into it to tackle the RNA inverse folding task. This method uses flow matching~\cite{watson2023novo} to iteratively refine RNA structures and conditions sequence prediction on the previously predicted conformation to approximate the true RNA sequence. By capturing RNA's flexibility and multiple conformations, this approach ensures that the predicted sequences are both nucleotide-accurate and reflective of RNA's dynamic nature. Therefore, when integrated with flow matching, our model iteratively samples the identity of each nucleotide based on the conditional probability distribution derived from the previously generated sequence, enhancing its ability to generate functional RNA sequences that align with desired structural properties.

\section{Experiments}

\begin{table*}[h]
%\small  % Reduce font size
%\setlength{\tabcolsep}{4pt}  % Adjust column separation
\caption{\textbf{Comparison results with gRNAde and its variant with hypergraph}. We reported Mean $\pm$ Standard Error of the Mean (SEM) for RMSD, lDDT and RMSD on two splits of PDBDind, and Validity, Diversity and Novelty on RNAsolo. }
\label{table:total}
\resizebox{\textwidth}{!}{
\begin{tabular}{l|ccc|ccc|ccc}
\toprule
\midrule
\multicolumn{1}{c|}{\multirow{2}{*}{Model}} & \multicolumn{3}{c|}{\textit{RF2NA Split for PDBBind}} & \multicolumn{3}{c|}{\textit{Sequence-Similar Split for PDBBind}} & \multicolumn{3}{c}{\textit{RNAsolo}} \\
\cmidrule{2-4} \cmidrule{5-7} \cmidrule{8-10}
 & RMSD($\downarrow$) & RNA Recovery($\uparrow$) & lDDT($\uparrow$) & RMSD($\downarrow$) & RNA Recovery($\uparrow$) & lDDT($\uparrow$) & Validity($\uparrow$) & Diversity($\uparrow$) & Novelty($\downarrow$) \\
\midrule
gRNAde & 13.51$\pm$1.26 & 0.28$\pm$0.08 & 0.51$\pm$0.02 & 17.86$\pm$1.02 & 0.29$\pm$0.02 & 0.56$\pm$0.01 & \textbf{0.27}$\pm$\textbf{0.011} & 0.43$\pm$0.005 & 0.57$\pm$0.009 \\
gRNAde + Hypergraph & \textbf{12.46}$\pm$\textbf{0.75} & 0.28$\pm$0.03 & 0.54$\pm$0.02 & 17.66$\pm$1.03 & 0.30$\pm$0.02 & 0.56$\pm$0.01 & 0.24$\pm$0.017 & 0.46$\pm$0.009 & 0.53$\pm$0.009 \\
\midrule
HyperRNA & 12.56$\pm$0.99 & \textbf{0.29}$\pm$\textbf{0.03} & \textbf{0.56}$\pm$\textbf{0.02} & \textbf{17.51}$\pm$\textbf{0.96} & \textbf{0.31}$\pm$\textbf{0.01} & \textbf{0.56}$\pm$\textbf{0.01} & 0.24$\pm$0.012 & \textbf{0.47}$\pm$\textbf{0.008} & \textbf{0.53}$\pm$\textbf{0.007} \\
\midrule
\bottomrule

\end{tabular}
}
\end{table*}

\subsection{Datasets}
We focused on two primary tasks to evaluate our HyperRNA model. First, we applied it as an inverse folding tool to predict protein-binding RNA sequences using the \textbf{PDBBind} dataset~\cite{pdbbind}, which contains 1192 RNA-protein complexes. We conducted experiments on two distinct data splits.  
The first split accounts for RF2NA pre-training~\cite{rf2na}, where all complexes present in the RF2NA validation or test set were assigned to the test set. The remaining complexes were randomly divided into training and validation sets in a 9:1 ratio. 
The second split was implemented based on RNA sequence similarity. RNA chains were clustered using CD-HIT~\cite{cd-hit}, and two sequences were divided into the same cluster if they share $\ge$ 80$\%$ sequence identity \cite{Joshi2023}. These clusters were then randomly partitioned into training, validation and test sets in an 8:1:1 ratio to further evaluate the model's capacity to generalize across a broad spectrum of RNA sequences. We further assessed the model’s ability to generalize across a wide range of RNA sequences using the \textbf{RNAsolo} dataset~\cite{Adamczyk2022}. RNA chains were clustered using CD-HIT~\cite{cd-hit}, and these clusters were then randomly divided into training, validation, and test sets in an 8:1:1 ratio for a total of 4223 RNA sequences.

\subsection{Implementation Details}
We conducted experiments on a workstation with an NVIDIA RTX 4090-24GB GPU using the PyTorch platform \cite{pytorch}. For the protein-binding RNA inverse folding task, HyperRNA served as a collaborative module within a broader RNA design framework \cite{rnaflow} for joint training. In our HyperRNA model, we set the number of multi-head attention layers to 3, with 3 encoder and 3 decoder layers, utilizing 128 scalar and 16 vector node features, along with 32 scalar and 1 vector edge features. A dropout rate of 0.1 was applied. The model was trained for 100 epochs using the Adam optimizer with an initial learning rate of 0.0001. Three key metrics were used: 
\begin{itemize}
    \item \textbf{Root Mean Squared Deviation (RMSD)} is computed between the predicted RNA structure and the ground-truth structure on all backbone atoms.
    \item \textbf{RNA recovery} refers to the proportion of RNA sequences correctly predicted for a given RNA backbone.
    \item \textbf{Local Distance Difference Test (lDDT)} evaluates the accuracy of predicted structures on $C\alpha$ atoms.
\end{itemize}

For the RNA sequences generation task, HyperRNA was pretrained~\cite{Adamczyk2022} to handle diverse RNA sequences. The model architecture remained unchanged, but the training epoch was reduced to 50 epochs. Three metrics are employed:
\begin{itemize}
    \item \textbf{Validity} is determined by whether the self-consistency TM-score (scTM) is $\ge$ 0.45. Each generated backbone is inverse folded, and $N_{\text{seq}} = 8$ generated sequences are passed into RhoFold~\cite{rhofold} to compute the scTM between the predicted RhoFold structure and the backbone at the $C4'$ level. 
    \item \textbf{Diversity} is the ratio of unique structural clusters to the total number of samples, where the unique clusters are computed using qTMclust~\cite{gtm} with TM-score $\ge$ 0.45. 
    \item \textbf{Novelty} measures the degree of similarity between the generated backbones and training samples. Among the valid samples, US-align~\cite{gtm} is used to assess structural differences at the $C4'$ level compared to the training set.
    % For every backbone, the highest TM-score between the backbone and training samples is assigned, and the average TM-score of the set of same length is referred to as pdbTM
\end{itemize}

Among these metrics, higher values in RNA recovery, lDDT, Validity and Diversity are preferred, while lower values in RMSD and novelty are considered better.

\subsection{Comparison and Discussion}
Since there are currently few models that address the 3D RNA inverse folding problem, we compared our model with gRNAde~\cite{RD8} and its variants that incorporate a hypergraph component to assess the impact of hypergraphs. To ensure a fair comparison, we trained all models on the same datasets and evaluated them using an identical set of RNA backbones, as shown in Table~\ref{table:total}.

\begin{figure}[t]
    \centering
    \includegraphics[width=0.94\linewidth]{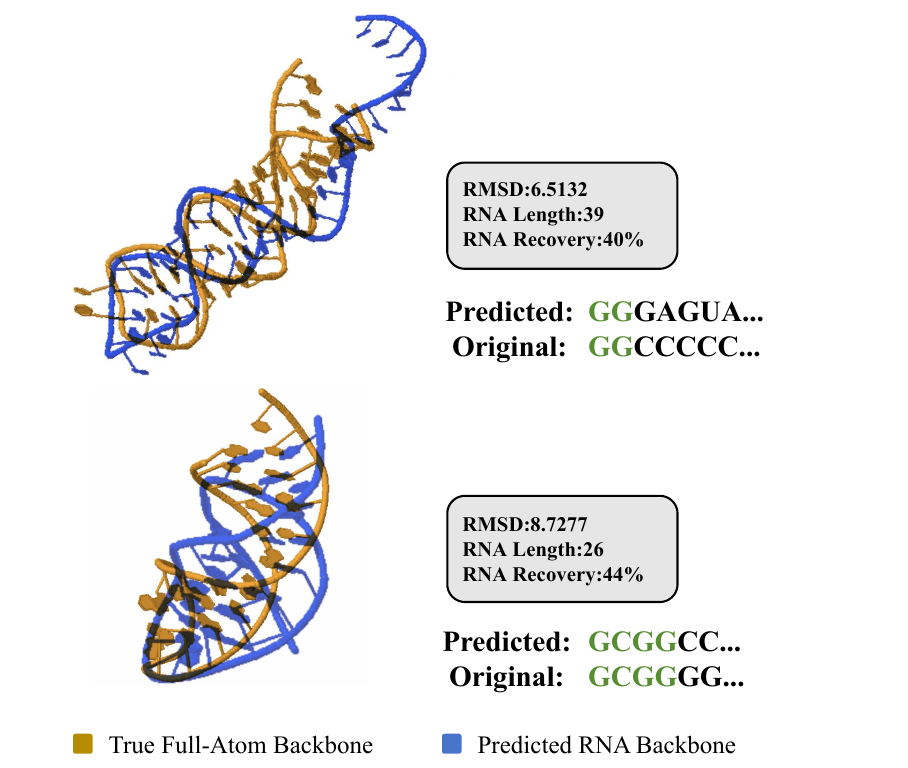}
    \caption{\textbf{Visualizations of RNA structures and sequences generated by HyperRNA}. Top: The crystal structure of Rev (PDB ID: 4PMI). Bottom: Design of RNA for composing the crystal structure of the \textit{Drosophila melanogaster} SNF (PDB ID: 6F4G).}
    \label{fig:RS}
\end{figure}

Our HyperRNA consistently excelled across multiple evaluation metrics on two datasets. On the \textit{RF2NA split} and \textit{Sequence-Similar split} from the PDBBind dataset, HyperRNA demonstrated superior RNA recovery rates, achieving scores of 0.29 and 0.31 respectively. These results highlight its enhanced ability to accurately recover RNA sequences that align with target structures. HyperRNA also attained the highest lDDT scores, with values of both 0.56, indicating its strong proficiency in predicting accurate RNA 3D structures. On the RNAsolo dataset, HyperRNA showed competitive performance, achieving the highest Diversity score of 0.47 and the lowest Novelty score of 0.53, reflecting its effectiveness in generating diverse and novel RNA backbone structures. Additionally, the integration of hypergraphs into gRNAde resulted in noticeable improvements in RMSD, RNA Recovery and lDDT scores, demonstrating the efficacy of hypergraph-based methods for enhanced structural modeling.

\subsection{Visualization}
We further conducted qualitative experiments to validate the effectiveness of our model on the \textit{RF2NA split} and \textit{the sequence-similar split}, as shown in Fig.~\ref{fig:RS}. For each dataset, we provided visualizations\footnote{\url{https://genesilico.pl/SimRNAweb/}} of both the original RNA structures and their corresponding structures generated through our inverse folding process, along with the measured RMSD and RNA recovery metrics. Specifically, we showcased the structure and sequence design of two RNA molecules: the component of the crystal structure of Rev (PDB ID: 4PMI) and the RNA designed for composing the crystal structure of \textit{Drosophila melanogaster} SNF (PDB ID: 6F4G). By analyzing the visual differences and consistencies between the original and generated structures, we demonstrated the robustness of our model in handling diverse RNA sequences and its potential to advance RNA design tasks.

\begin{figure}[t]
    \centering
    \includegraphics[width=1\linewidth]{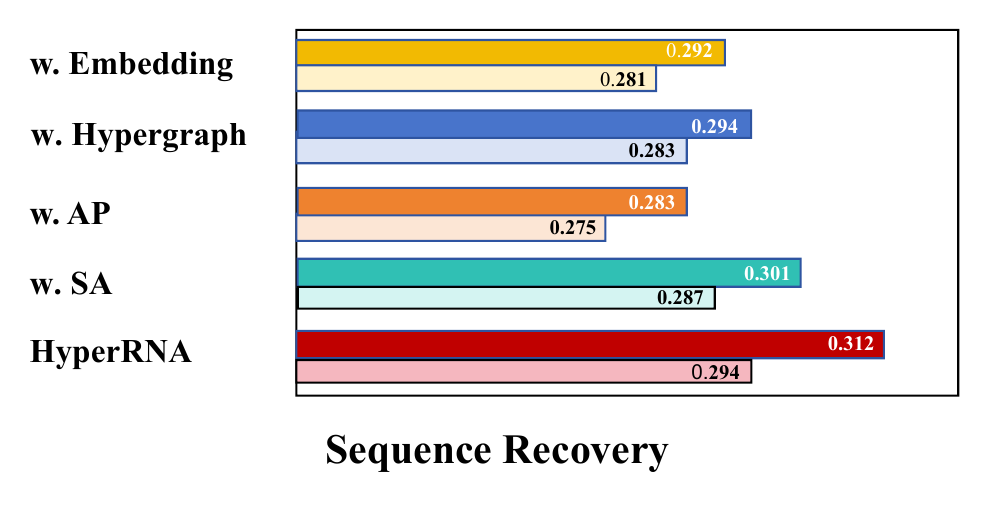}
    \caption{\textbf{Ablation Studies of our HyperRNA on RNA sequence recovery}. The top side represents the result on the \textit{sequence-similar split}, while the bottom side shows the result on the \textit{RF2NA split}. ``AP'' refers to attention pooling, while ``SA'' refers to self-attention embedding.}
    \label{fig:ablation}
\end{figure}

% \usepackage{multirow}

\subsection{Ablation Studies}

We conducted extensive experiments to verify the effectiveness of each component in our proposed HyperRNA model. We reported the RNA recovery and RMSD metrics on the \textit{RF2NA split} and the \textit{sequence-similar split} of the PDBBind dataset, as shown in Fig.~\ref{fig:ablation} and Table~\ref{RMSD} respectively.

Compared with HyperRNA, using only the hypergraph-based encoder led to a significant decline in performance, with RMSD increasing by 1.0 and 0.4 for the \textit{sequence-similar split} and \textit{RF2NA split} respectively, and RNA recovery decreasing by 0.020 and 0.013. This highlights the effectiveness of hypergraphs in capturing complex RNA relationships. When retaining the hypergraph-based encoder without the attention embedding module, we observed a reduction in sequence reconstruction accuracy, with RNA recovery decreasing by 0.018 and 0.011 for the \textit{sequence-similar split} and \textit{RF2NA split}, respectively. This demonstrates the effectiveness of the attention embedding in capturing essential structural dependencies within RNA, which are crucial for accurate sequence prediction and for maintaining structural fidelity.

To further assess the specific contributions of processing scalar and vector features within the attention embedding module, we applied the self-attention exclusively to vector features. This led to a significant drop in performance, with RNA recovery decreasing by 0.011 and 0.007 in two splits and RMSD substantially increasing to 18.0 in the \textit{sequence-similar split}. This finding underscores the importance of scalar features in modeling spatial dependencies for accurate RNA structure prediction. Conversely, when using only attention pooling on scalar features, the preservation of RNA secondary structure was further degraded, with RNA recovery decreasing from 0.312 and 0.294 to 0.283 and 0.275. This demonstrates the necessity of integrating both feature types. These results emphasize the importance of a comprehensive attention strategy that combines both vector and scalar features to achieve optimal performance, particularly in the \textit{sequence-similar split}.

\begin{table}[t]
\caption{\textbf{Ablation Studies our HyperRNA on the PDBBind Dataset}: ``AP'' refers to attention pooling for scalar features, while ``SA'' refers to self-attention embedding for vector features. }
\centering
\label{RMSD}
\begin{tabular}{l|cc}
\toprule
\midrule
\multicolumn{1}{c|}{\multirow{2}{*}{Module}} & \multicolumn{2}{c}{RMSD ($\downarrow$)}           \\
\cmidrule{2-3}
& \textit{RF2NA Split} & \textit{Sequence-Similar Split} \\
\midrule
\textbf{w.} Embedding                                        & \multicolumn{1}{c|}{13.5$\pm$1.06}        & 17.9$\pm$1.06                   \\
\textbf{w.} Hypergraph                                        & \multicolumn{1}{c|}{12.5$\pm$1.29}        & 17.7$\pm$1.07                   \\
AP+Hypergraph                                          & \multicolumn{1}{c|}{12.6$\pm$1.25}        & 17.6$\pm$0.95                   \\
SA+Hypergraph                                          & \multicolumn{1}{c|}{13.0$\pm$1.77}        & 18.0$\pm$1.00                    \\
\midrule
HyperRNA                                               & \multicolumn{1}{c|}{\textbf{12.5}$\pm$\textbf{0.99}}        & \textbf{17.5}$\pm$\textbf{0.96}   \\
\midrule
\bottomrule
\end{tabular}

\end{table}

\section{Conclusion}

In this paper, we propose a generative model called HyperRNA to address the 3D RNA inverse folding problem. Our approach leverages the power of hypergraphs to enhance traditional graph-based methods.
HyperRNA comprises three key steps: \textit{preprocessing}, which constructs graph structures from given RNA and protein inputs; \textit{encoding}, which captures the intricate associations within these graphs and \textit{decoding}, which employs an autoregressive model to generate the sequence. Through extensive evaluations on the PDBBind and RNAsolo datasets, HyperRNA demonstrated superior performance across several metrics, including RNA recovery, structural accuracy and the ability to generate diverse and novel RNA structures. Our results consistently showed that HyperRNA not only outperforms existing methods but also significantly benefits from the integration of hypergraphs. These findings highlight the potential of hypergraph-based methods in RNA design, providing a robust framework for more accurate and diverse RNA structure modeling. Future work will focus on further optimizations and exploring the applications of HyperRNA across a broader range of RNA-related tasks.

\bibliographystyle{IEEEtran} 
%\biboptions{authoryear}
\bibliography{bibfile}

\end{document}